\documentclass[twocolumn]{article}
\usepackage{dx}
\usepackage{times}
\usepackage{graphicx}
\usepackage{algorithm}
\usepackage{algorithmic}
\usepackage{rotating}
\usepackage{algo}
\usepackage{amssymb}
\usepackage{xyling}

\begin{document}
\dxHeader{
  \title{\textsc{CoreDiag}: Eliminating Redundancy in Constraint Sets}
  \author{ 			Alexander Felfernig 	\affiliationNumber{1}
					\and 	Christoph Zehentner	\affiliationNumber{1} \and 	Paul Blazek	\affiliationNumber{2}
	}
  \address{
					\affiliation{1}
					            {Graz University of Technology, Inffeldgasse 16b, 8010 Graz, Austria}
					            {alexander.felfernig@ist.tugraz.at \\
					             christoph.zehentner@ist.tugraz.at}
					                
\affiliation{2}
					            {cyLEDGE Media GmbH, Schottenfeldgasse 60, 1070 Vienna, Austria}
					            {p.blazek@cyledge.com} 
}

	\maketitle
  \begin{abstract}
Constraint-based environments such as configuration systems, recommender systems, and scheduling systems support users in different decision making scenarios. These environments exploit a knowledge base for determining solutions of interest for the user. The development and maintenance of such knowledge bases is an extremely time-consuming and error-prone task. Users often specify constraints which do not reflect the real-world. For example,  redundant constraints are specified which often increase both, the effort for calculating a solution and efforts related to knowledge base development and maintenance. In this paper we present a new algorithm (\textsc{CoreDiag}) which can be exploited for the determination of minimal cores (minimal non-redundant constraint sets). The algorithm is especially useful for distributed knowledge engineering scenarios where the degree of redundancy can become  high. In order to show the applicability of our approach, we present an empirical study conducted with commercial configuration knowledge bases.
  \end{abstract}

}

\section{Introduction}

The central element of a constraint-based application is a knowledge base (constraint set). When developing and maintaining  constraint sets, users are often defining \emph{faulty} constraints (the system calculates solutions which are not allowed or -- in the worst case -- no solution can be found) \cite{paperBakker1993,paperfelfernig2004} or \emph{redundant} constraints which are not needed to express the  domain knowledge in a complete fashion  \cite{paperfreuder1995,paperPiette2008,paperFahad1993,paperLevy1993}. In this paper we focus on situations where users are defining redundant constraints which -- when deleted from the constraint set (knowledge base) -- do not change the semantics of the remaining constraint set. More formally, if C=\{$c_1, c_2, ..., c_n$\} is the initial set of constraints defined for the  knowledge base and one constraint $c_i$ is redundant, then $(C-\{c_i\})$ $\cup ~ \overline{C}$ is inconsistent ($\overline{C}$ is the negation of C).

Redundancy elimination in knowledge bases is a topic extensively investigated by AI research. The identification of redundant constraints plays a major role, for example,  in the development and maintenance of configuration knowledge bases (see, e.g., \cite{paperfreuder1995}). The authors introduce concepts for the detection of redundant constraints in conditional constraint satisfaction problems (CCSPs). The  approach is based on the idea of analyzing the solution space of the given problem (on the level of individual solutions) in order to detect different types of redundant constraints. \cite{paperPiette2008} provide an in-depth discussion of the role of redundancy elimination in SAT solving. They introduce an (incomplete) algorithm for the elimination of redundant clauses and show its applicability on the basis of an empirical study. The role of redundancies in ontology development is  analyzed by \cite{paperFahad1993}. The authors  point out the importance of redundancy elimination and discuss typical modeling errors that occur during ontology development and maintenance. \cite{paperGrimm2011} introduce algorithms for redundancy elimination in OWL ontologies. The authors propose an algorithm that computes redundant axioms by exploiting prior knowledge of the dispensibility of axioms. \cite{paperLevy1993} analyze two types of redundancies in Datalog programs. First, they interpret redundancy in terms of \emph{reachability}, i.e., rules and predicates are eliminated that are \emph{not} part of any derivation tree. Second, redundancy is defined on the basis of the concepts of \emph{minimal derivation trees} which do not include any pair of identical atoms where one is the predecessor of the other one. 

All the mentioned approaches focus on the identification of redundant constraints in centralized scenarios where a knowledge engineer is interested in identifying redundant constraints in the given knowledge base. In such scenarios it is assumed that only a small subset of the given constraints is redundant (this assumption is also denoted as \emph{low redundancy assumption} \cite{paperGrimm2011}). Existing algorithms are focusing on such centralized scenarios. In this paper we go one step further and propose an algorithm which is especially useful in distributed knowledge engineering scenarios where we can expect a \emph{larger number of redundant constraints} due to the fact that different contributors add constraints which are related to the same topic (see, e.g., \cite{paperchklovski2005,paperRichardson2003}) -- we denote the assumption of larger sets of redundant constraints the \emph{high redundancy assumption}. For example, we envision a scenario where a large number of users propose constraints to be applied by a constraint-based configuration or recommendation engine \cite{paperfelfernigburke2008} and the task of an underlying diagnosis algorithm is to identify minimal sets of constraints which retain the semantics of the original constraint set -- we denote such constraint sets as \emph{minimal cores}. Note that the following discussions are based on the assumption of \emph{consistent constraint sets}. Methods for consistency restoration are discussed in \cite{paperBakker1993,paperfelfernig2004,FriedrichISWC2005,paperfelfernig2011}.

The major contributions of our  paper are the following. First, we introduce a new algorithm which allows for a more efficient determination of redundant constraints especially in the context of distributed (community-based) knowledge engineering scenarios. Second, we present the results of a performance analysis of our algorithm conducted with real-world configuration knowledge bases.

The remainder of this paper is organized as follows. In Section 2 we introduce a simple example configuration knowledge base from the automotive domain. In Section 3 we introduce a basic algorithm for the determination of redundant constraints in centralized settings (\textsc{Sequential}). In Section 4 we introduce the \textsc{CoreDiag} algorithm. Thereafter we report the results of a performance evaluation conducted with  real-world configuration knowledge bases (Section 5). The paper is concluded with Section 6.

\section{Working Example}
For illustration purposes we use a car configuration knowledge base throughout this paper. A configuration task can be defined as a basic constraint satisfaction problem (CSP) \cite{booktsang1993} (see the following definition).\footnote{Note that the presented concepts are as well applicable to other types of knowledge representations such as SAT or description logics.}

\paragraph {Definition (Configuration Task)}: A configuration task can be defined as a CSP (V, D, C). V = \{$v_1, v_2, ..., v_n$\} is a set of finite domain variables. D = \{$dom(v_1), dom(v_2), ..., dom(v_n)$\} is a set of corresponding domain definitions where dom($v_k$) is the domain of the variable $v_k$. C = $C_{KB}~ \cup ~C_R$ where $C_{KB}~=~$ \{$c_1, c_2, ..., c_q$\} is a set of domain-specific constraints (the configuration knowledge base) and $C_R~=~\{c_{q+1}, c_{q+2}, ..., c_t\}$ is a set of customer requirements (as well represented as constraints). 

The following configuration task will be used as a working example throughout the paper. The variable \emph{type} represents the type of the car, \emph{pdc} is the park distance control feature, \emph{fuel} represents the average fuel consumption per 100 kilometers, a \emph{skibag} allows  convenient ski stowage inside the car, and \emph{4-wheel} represents the actuation type (4-wheel supported or not supported). These variables represent the possible combinations of customer requirements. The set $C_{KB}~=~\{c_1, c_2, c_3, c_4, c_5\}$ defines additional restrictions on the set of possible customer requirements $C_R~=~\{c_6, c_7, c_8, c_9, c_{10}\}$.
\begin{itemize}
\item{$V=\{type, fuel, skibag, 4-wheel, pdc\}$}
\item{$D=\{$\\
$~~~dom(type)=\{city, limo, combi, xdrive\}$,\\ 
$~~~dom(fuel)=\{4l, 6l, 10l\}$,\\
$~~~dom(skibag)=\{yes, no\}$,\\
$~~~dom(4-wheel)=\{yes, no\}$,\\
$~~~dom(pdc)=\{yes, no\}\}$}
\item{$C_{KB}~=~\{$\\
$~~~c_1: 4-wheel = yes \rightarrow type = xdrive$,\\
$~~~c_2: skibag = yes \rightarrow type \neq city$,\\
$~~~c_3: fuel = 4l \rightarrow type = city$,\\
$~~~c_4: fuel = 6l \rightarrow type \neq xdrive$,\\
$~~~c_5: type = city \rightarrow fuel \neq 10l$\}
\item{$C_R~=~\{$\\
$~~~c_6: 4-wheel=no,$\\
$~~~c_7: fuel=4l,$\\
$~~~c_8: type=city,$\\
$~~~c_9: skibag=no,$\\
$~~~c_{10}: pdc=yes$}\}}
\end{itemize}

On the basis of this example configuration task we now give a definition of a corresponding configuration (solution).

\paragraph {Definition (Configuration)}: A configuration (solution) for a configuration task is an instantiation I=$\{v_1=ins_1, v_2=ins_2, ..., v_n=ins_n\}$ where $ins_k$ is an element of the domain of $v_k$. A configuration is \emph{consistent} if the assignments in $I$ are consistent with the constraints in $C$. A \emph{complete} solution is one in which all the variables are instantiated. Finally, a configuration is \emph{valid} if it is both, consistent and complete.

A configuration for our example configuration task would be $I~=~\{4-wheel=no, fuel=4l, type=city, skibag=no,  pdc=yes\}$.

\section{Determining Redundant Constraints}
Let us now consider a simple adaptation of the original set of constraints $C_{KB}$ which we denote with $C_{KB}'$. $C_{KB}'$ includes an additional constraint $c_a$ which has been added by a knowledge engineer.

$C_{KB}'~=~\{$\\
$~~~~~~~~~c_a: skibag \neq no \rightarrow type = limo ~\vee$\\ 
$~~~~~~~~~~~~~~~~type = combi ~\vee$\\
$~~~~~~~~~~~~~~~~type = xdrive$,\\
$~~~~~~~~~c_1: 4-wheel = yes \rightarrow type = xdrive$,\\
$~~~~~~~~~c_2: skibag = yes \rightarrow type \neq city$,\\
$~~~~~~~~~c_3: fuel = 4l \rightarrow type = city$,\\
$~~~~~~~~~c_4: fuel = 6l \rightarrow type \neq xdrive$,\\
$~~~~~~~~~c_5: type = city \rightarrow fuel \neq 10l$\}

It is obvious that $c_a$ is redundant since it does not further restrict the solution space defined by the constraints $C_{KB}~=~\{c_1, c_2, c_3, c_4, c_5\}$.  In order to discuss constraint \emph{redundancy} on a more formal level, we introduce the following definitions.

\paragraph {Definition (Redundant Constraint)}: Let $c_a$ be a constraint element of the configuration knowledge base $C_{KB}$. $c_a$ is called  redundant  \emph{iff} $C_{KB}~-~\{c_a\} \models c_a$. If this condition is not fulfilled, $c_a$ is said to be \emph{non-redundant}. Redundancy can also be analyzed by checking $C_{KB}~-~\{c_a\} ~\cup~ \overline{C_{KB}}$ for consistency -- if consistency is given, $c_a$ is non-redundant.

Iterating over each constraint of $C_{KB}$, executing the non-redundancy check $C_{KB}~-~\{c_a\}~ \cup ~ \overline{C_{KB}}$, and deleting redundant constraints from $C_{KB}$ results in a set of non-redundant constraints (the \emph{minimal core}). If the non-redundancy check fails (no solution can be found), the constraint $c_a$ is redundant and can be deleted from $C_{KB}$. Otherwise (the non-redundancy check is successful), $c_a$ is non-redundant.

\paragraph {Definition (Minimal Core)}: Let $C_{KB}$ be a configuration knowledge base. $C_{KB}$ is denoted as \emph{minimal core} \emph{iff} $\forall c_i \in C_{KB}: C_{KB}-\{c_i\} ~\cup ~ \overline{C_{KB}}$ is consistent. Obviously, $C_{KB}~ \cup ~ \overline{C_{KB}} \models \bot$.

The principle of the following algorithm (\emph{\textsc{Sequential}} -  Algorithm 1) is often used for determining such redundancies (see, e.g., \cite{paperPiette2008,paperGrimm2011}). 

\algsetup{ linenosize={\small , linenodelimiter=. } }%
\begin{algorithm}[ht]
 \caption{\textsc{Sequential}($C_{KB}$)$:\Delta$}
\label{alg:Sequential} \begin{algorithmic} 

\STATE \COMMENT{$C_{KB}$:  configuration knowledge base} 
\STATE \COMMENT{$\overline{C_{KB}}$:  the complement of $C_{KB}$}  
\STATE \COMMENT{$\Delta$: set of redundant constraints} 
\STATE $C_{KBt} \gets C_{KB};$
\FORALL {$c_i$ $in$ $C_{KBt}$} 
\IF {$isInconsistent((C_{KBt}-c_i) ~ \cup ~ \overline{C_{KB}})$} \STATE $C_{KBt} ~ \gets$ $C_{KBt} ~ - ~ \{c_i\};$ \ENDIF
\ENDFOR
\STATE $\Delta \gets C_{KB} ~ - ~ C_{KBt};$ 
\STATE $return ~ \Delta;$ 
\end{algorithmic} 
\end{algorithm}

The approach of \emph{\textsc{Sequential}} is straightforward: each individual constraint $c_i$ is evaluated w.r.t. redundancy by checking whether  $C_{KBt}~-~ c_i$ is still \emph{inconsistent} with $\overline{C_{KB}}$ . If this is the case, $c_i$ can be considered as redundant. If $C_{KBt}~-~ c_i$ is \emph{consistent} with $\overline{C_{KB}}$, $c_i$ is a non-redundant constraint since its deletion induces consistency with $\overline{C_{KB}}$. Applying the algorithm \emph{\textsc{Sequential}} to our example $C_{KB}'$ results in $\Delta$ = $\{c_a\}$ since $C_{KB}-\{c_a\}~\cup~\overline{C_{KB}}$ is inconsistent and no further constraint $c_i$ can be deleted from $C_{KB}$ such that $C_{KB}~-~\{c_a\}~-~\{c_i\}$ is still inconsistent.

The problem of checking whether a given constraint can be inferred from the remaining part of a constraint set has  shown to be Co-NP-complete in the general case \cite{paperPiette2008}. The major goal of our work was to figure out whether there exist alternative algorithms that have a better runtime performance compared to \emph{\textsc{Sequential}} in situations with a large amount of redundant constraints in $C_{KB}$. Large  amounts of redundant constraints typically occur in distributed knowledge engineering scenarios where a large number of users specify rules that in the following have to be aggregated into one consistent constraint set (see, e.g., \cite{paperchklovski2005}).

In the following section we introduce the \emph{\textsc{CoreDiag}} algorithm which is a valuable alternative to \emph{\textsc{Sequential}} in situations with a large number of redundant constraints. After having introduced \emph{\textsc{CoreDiag}} we will analyze the performance of both algorithms (\emph{\textsc{Sequential}} and \emph{\textsc{CoreDiag}}) on the basis of real-world configuration knowledge bases (Section 5).

\section{CoreDiag}

The \emph{\textsc{CoreDiag}} algorithm (together with \emph{\textsc{CoreD}}) is based on the principle of divide-and-conquer: whenever a set $S$ which is a subset of $C_{KB}$ is inconsistent with $\overline{C_{KB}}$, it is or contains a minimal core, i.e. a set of constraints which preserve the semantics of $C_{KB}$. In our implementation \emph{\textsc{CoreD}} is responsible for determining such minimal cores, \emph{\textsc{CoreDiag}} returns the complement of a minimal core which is a maximal set of redundant constraints in $C_{KB}$. \emph{\textsc{CoreD}} is based on the principle of \emph{QuickXPlain} \cite{paperjunker2004} -- as a consequence a minimal core (minimal set of constraints that preserve the semantics of $C_{KB}$) can be interpreted as  a minimal conflict, i.e., a minimal set of constraints that are inconsistent with $\overline{C_{KB}}$.

\emph{\textsc{CoreD}} allows the determination of \emph{preferred} minimal cores since the algorithm is based on the assumption of a strict lexicographical ordering of the constraints in $C_{KB}$. On an informal level a preferred minimal core can be characterized as follows: if we have different options for choosing a minimal core, we would select the one with the most agreed-upon constraints. For more details on the role of strict lexicographical orderings of constraints we  refer the reader to the work of \cite{paperjunker2004} and \cite{paperfelfernig2011}. 

The \emph{\textsc{CoreDiag}} algorithm generates $\overline{C_{KB}}$ from $C_{KB}$. It then activates \emph{\textsc{CoreD}} (see Algorithm 3) which determines a minimal core on the basis of a divide-and-conquer strategy that divides the constraints in $C$ into two subsets ($C_1$ and $C_2$) with the goal to figure out whether one of those subsets already contains a minimal core. If $C_2$ contains a minimal core, $C_1$ is not further taken into account. If $C$ contains only one element ($c_\alpha$) and $B$ is still consistent, then $c_\alpha$ is part of the minimal core.

\algsetup{ linenosize={\small , linenodelimiter=. } }%
\begin{algorithm}[ht]
 \caption{\textsc{CoreDiag} ($C_{KB}$)$:\Delta$}
\label{alg:CoreDiag} \begin{algorithmic} 

\STATE \COMMENT{$C_{KB}~=~\{c_1, c_2, ..., c_n\}$} 
\STATE \COMMENT{$\overline{C_{KB}}$:  the complement of $C_{KB}$}  
\STATE \COMMENT{$\Delta$: set of redundant constraints} 

\STATE $\overline{C_{KB}} \gets \{\neg c_1 \vee \neg c_2 \vee ... \vee \neg c_n\};$
\STATE $return (C_{KB}~-~ \textsc{CoreD}(\overline{C_{KB}}, \overline{C_{KB}}, C_{KB}));$ 

\end{algorithmic} 
\end{algorithm}

\algsetup{ linenosize={\small , linenodelimiter=. } }%
\begin{algorithm}[ht]
 \caption{\textsc{CoreD}($B, D, C=\{c_1, c_2, ..., c_p\}$)$:\Delta$}
\label{alg:CoreD} \begin{algorithmic} 

\STATE \COMMENT{B: consideration set} 
\STATE \COMMENT{D: constraints added to B}  
\STATE \COMMENT{C: set of constraints to be checked for redundancy} 

\IF {$D \neq \emptyset ~ and ~ inconsistent(B)$} \STATE $return ~ \emptyset;$ \ENDIF
\IF {$singleton(C)$} \STATE $return (C);$ \ENDIF
\STATE $k \gets \lceil \frac{r}{2} \rceil;$
\STATE $C_1 \gets \{c_1, c_2, ..., c_k\};$
\STATE $C_2 \gets \{c_{k+1}, c_{k+2}, ..., c_p\};$
\STATE $\Delta_1 \gets \textsc{CoreD}(B~\cup~C_2, C_2, C_1);$
\STATE $\Delta_2 \gets \textsc{CoreD}(B~\cup~\Delta_1, \Delta_1, C_2);$
\STATE $return (\Delta_1 ~\cup ~ \Delta_2);$
\end{algorithmic} 
\end{algorithm}

\section{Evaluation}

\begin{table*}[ht]
\center
\begin{tabular}{|l|c|c|c|c|c|}
\hline
  
    & \multicolumn{5}{|c|}{Redundancy Rate} \\
    KB ($|c_{KB}|$) & Alg.  & \textasciitilde 0-10\% &  \textasciitilde 50\% & \textasciitilde 75\% &  \textasciitilde 87.5\% \\
    \hline
Bike\_A (32) & S & 32.0 / 205.4 / 0 & 64.0 / 408.6 / 32 & 128.0 / 1209.0 / 96 & 256.0 / 4073.2 / 224 \\ \hline
Bike\_A (32) & CD & 63.0 / 614.4 / 0 & 88.8 / 863.2 / 32 & 106.6 / 1352.0 / 96 & 107.4 / 1737.2 / 224 \\ \hline
Bike\_B  (35) & S & 35.0 / 256.8 / 1 & 70.0 / 616.4 / 36 & 140.0 / 1710.0 / 106 & 280.0 / 4854.0 / 246 \\ \hline
Bike\_B (35) & CD & 68.6 / 693.4 / 1 & 94.0 / 960.8 / 36 & 109.6 / 1365.2 / 106 & 117.8 / 1893.0 / 246 \\ \hline
Bike\_C (37) & S & 37.0 / 297.0 / 1 & 74.0 / 696.6 / 38 & 148.0 / 1824.8 / 112 & 296.0 / 5722.8 / 260 \\ \hline
Bike\_C (37) & CD & 72.4 / 703.6 / 1 & 101.2 / 1091.2 / 38 & 114.8 / 1524.2 / 112 & 122.2 / 2115.4 / 260 \\ \hline
Bike\_D (34) & S & 34.0 / 280.2 / 1 & 68.0 / 606.8 / 35 & 136.0 / 1672.0 / 103 & 272.0 / 5033.6 / 239 \\ \hline
Bike\_D (34) & CD & 66.2 / 727.6 / 1 & 94.8 / 1031.8 / 35 & 104.8 / 1433.8 / 103 & 114.8 / 2000.8 / 239 \\ \hline
Bike\_E (35) & S & 35.0 / 254.2 / 9 & 70.0 / 601.0 / 44 & 140.0 / 1628.6 / 114 & 280.0 / 5124.8 / 254 \\ \hline
Bike\_E (35)& CD & 60.8 / 663.0 / 9 & 83.4 / 821.4 / 44 & 96.0 / 1182.6 / 114 & 103.6 / 1671.2 / 254 \\ \hline
Bike\_F (33) & S & 33.0 / 274.0 / 1 & 66.0 / 601.8 / 34 & 132.0 / 1573.8 / 100 & 264.0 / 4525.0 / 232 \\ \hline
Bike\_F (33) & CD & 64.6 / 632.8 / 1 & 88.6 / 931.2 / 34 & 108.2 / 1345.6 / 100 & 110.8 / 1822.2 / 232 \\ \hline
Bike\_G (36) & S & 36.0 / 281.4 / 2 & 72.0 / 660.6 / 38 & 144.0 / 1729.8 / 110 & 288.0 / 5434.4 / 254 \\ \hline
Bike\_G (36) & CD & 70.6 / 714.6 / 2 & 96.0 / 939.8 / 38 & 111.6 / 1409.6 / 110 & 122.4 / 2081.8 / 254 \\ \hline
Bike\_H (24) & S & 24.0 / 194.2 / 0 & 48.0 / 398.8 / 24 & 96.0 / 1047.0 / 72 & 192.0 / 3010.0 / 168 \\ \hline
Bike\_H (24) & CD & 47.0 / 443.4 / 0 & 63.0 / 587.4 / 24 & 77.2 / 869.8 / 72 & 80.0 / 1240.4 / 168 \\ \hline
Bike\_I (35) & S & 35.0 / 268.4 / 1 & 70.0 / 647.0 / 36 & 140.0 / 1696.4 / 106 & 280.0 / 4976.4 / 246 \\ \hline
Bike\_I (35) & CD & 68.4 / 708.6 / 1 & 93.6 / 985.0 / 36 & 112.2 / 1371.4 / 106 & 117.0 / 1897.0 / 246 \\ \hline
Bike\_J (46) & S & 46.0 / 366.8 / 4 & 92.0 / 867.8 / 50 & 184.0 / 2309.8 / 142 & 368.0 / 7234.4 / 326 \\ \hline
Bike\_J (46) & CD & 88.4 / 896.0 / 4 & 119.4 / 1258.8 / 50 & 139.8 / 1886.8 / 142 & 142.0 / 2413.6 / 326 \\ \hline
Bike\_K (35) & S & 35.0 / 254.0 / 1 & 70.0 / 805.4 / 36 & 140.0 / 1852.8 / 106 & 280.0 / 5146.8 / 246 \\ \hline
Bike\_K (35) & CD & 68.8 / 712.4 / 1 & 95.6 / 1021.6 / 36 & 108.8 / 1374.2 / 106 & 117.6 / 1945.4 / 246 \\ \hline
Bike\_L (37) & S & 37.0 / 290.0 / 2 & 74.0 / 645.8 / 39 & 148.0 / 1822.0 / 113 & 296.0 / 5740.8 / 261 \\ \hline
Bike\_L (37) & CD & 71.4 / 716.4 / 2 & 96.6 / 1001.6 / 39 & 113.2 / 1425.8 / 113 & 111.0 / 1829.8 / 261 \\ \hline
Bike\_2 (32) & S & 32.0 / 883.0 / 3 & 64.0 / 2386.4 / 35 & 128.0 / 8218.8 / 99 & 256.0 / 37784.4 / 227 \\ \hline
Bike\_2 (32) & CD & 61.2 / 2165.2 / 3 & 85.4 / 3749.6 / 35 & 97.2 / 5693.2 / 99 & 108.0 / 10276.8 / 227 \\ \hline
esvs (21) & S & 21.0 / 340.0 / 0 & 42.0 / 870.8 / 21 & 84.0 / 2771.8 / 63 & 168.0 / 10231.8 / 147 \\ \hline
esvs (21) & CD & 41.0 / 724.0 / 0 & 56.0 / 1170.6 / 21 & 65.6 / 1844.0 / 63 & 71.0 / 3296.4 / 147 \\ \hline
fs (16) & S & 16.0 / 291.6 / 1 & 32.0 / 664.0 / 17 & 64.0 / 1989.2 / 49 & 128.0 / 7238.0 / 113 \\ \hline
fs (16) & CD & 30.6 / 658.8 / 1 & 42.0 / 933.4 / 17 & 49.2 / 1504.2 / 49 & 52.2 / 2431.8 / 113 \\ \hline
hypo (21) & S & 21.0 / 116.6 / 1 & 42.0 / 321.0 / 22 & 84.0 / 975.4 / 64 & 168.0 / 3297.6 / 148 \\ \hline
hypo (21) & CD & 40.6 / 383.8 / 1 & 55.2 / 549.0 / 22 & 62.2 / 802.2 / 64 & 71.0 / 1293.0 / 148 \\ \hline
large2 (185) & S & 130.0 / 2552.8 / 75 & 260.0 / 4721.8 / 260 & 520.0 / 7860.0 / 445 & 1040.0 / 15025.4 / 630 \\ \hline
large2 (185) & CD & 76.8 / 1868.8 / 75 & 79.8 / 2085.6 / 260 & 96.6 / 2834.0 / 445 & 103.0 / 3870.4 / 630 \\ \hline
\end{tabular}
\caption{Application of \textsc{Sequential (S)} (Algorithm 1) and \textsc{CoreDiag} (C) (Algorithm 2) to  configuration knowledge bases of www.itu.dk/research/cla/externals/clib. "Bikex": bicycles; "esvs": corporate networks; "fs": financial services (insurances); "hypo": financial services (investments); "large2": electronic circuits. Evaluation data: (\emph{\#TP-calls} / \emph{runtime} (\emph{ms}) / \emph{\#redundant constraints}).}
\label{label}
\end{table*}

\begin{table*} [ht]
\center
\begin{tabular}{|l|c|c|c|c|c|}
\hline
  
    & \multicolumn{5}{|c|}{Redundancy Rate} \\
    KB & redundancy-free  & \textasciitilde 0-10\% & \textasciitilde 50\% & \textasciitilde 75\% & \textasciitilde 87.5\% \\
    \hline

Bike\_A 	&	9.9   &   10.2   &   14.2   &   33.7   &   43.0    \\ \hline 
Bike\_B 	&	9.3   &   9.5   &   29.0   &   24.6   &   41.9    \\ \hline 
Bike\_C 	&	8.4   &   11.0   &   18.3   &   32.3   &   44.0    \\ \hline 
Bike\_D 	&	7.8   &   8.9   &   25.6   &   28.0   &   42.6    \\ \hline 
Bike\_E 	&	6.3   &   13.0   &   19.0   &   24.2   &   41.5    \\ \hline 
Bike\_F 	&	7.7   &   12.8   &   19.5   &   22.3   &   41.4    \\ \hline 
Bike\_G 	&	7.3   &   10.3   &   16.9   &   26.0   &   44.8    \\ \hline 
Bike\_H 	&	10.5   &   11.0   &   13.6   &   24.1   &   37.2    \\ \hline 
Bike\_I 	&	8.0   &   9.0   &   15.7   &   31.5   &   45.4    \\ \hline 
Bike\_J 	&	9.1   &   19.3   &   24.8   &   26.6   &   49.2    \\ \hline 
Bike\_K 	&	7.7   &   11.4   &   17.3   &   24.8   &   43.5    \\ \hline 
Bike\_L 	&	9.3   &   13.4   &   25.8   &   26.7   &   45.5    \\ \hline 
Bike\_2 	&	44.8   &   48.3   &   84.1   &   162.0   &   323.4    \\ \hline 
esvs 	&	22.7   &   26.0   &   45.1   &   79.5   &   157.8    \\ \hline 
fs 	&	22.6   &   24.2   &   44.0   &   76.5   &   149.7    \\ \hline 
hypo 	&	8.3   &   8.4   &   19.1   &   26.0   &   47.6    \\ \hline 
large2 	&	15.5   &   16.6   &   22.0   &   25.5   &   36.4   \\ \hline

\end{tabular}
\caption{Time in \emph{ms} needed for calculating a solution for a given configuration knowledge base version.}
\label{label}
\end{table*}

We now compare the performance of \emph{\textsc{CoreDiag}} with the \textsc{Sequential} algorithm discussed in Section 3. The worst case complexity (and best case complexity) of \emph{\textsc{Sequential}} in terms of the number of needed consistency checks is $n$ (the number of constraints in $C_{KB}$). Worst case and best case complexity are identical since \emph{\textsc{Sequential}} checks the redundancy of each individual constraint $c_i$ with respect to $C_{KB}$. In contrast, the worst case complexity of \emph{\textsc{CoreDiag}}  depends on the number of redundant constraints  in $C_{KB}$. The worst case complexity of \emph{\textsc{CoreDiag}} in terms of the number of needed consistency checks is  $2c*log_2(\frac{n}{c})+2c$ where $n$ is the number of constraints in $C_{KB}$ and $c$ is the minimal core size. The best case complexity in terms of the number of needed consistency checks can be achieved if all constraints element of the minimal core are positioned in one branch of the \emph{\textsc{CoreD}} search tree: $log_2(\frac{n}{c})+2c$. Consequently, the performance of \emph{\textsc{CoreDiag}} heavily relies on the number of constraints contained in the minimal core (the lower the number of constraints in the minimal core, the better the performance of \emph{\textsc{CoreDiag}}).

Table 1 reflects the results of our analysis conducted with the knowledge bases of the configuration benchmark.\footnote{www.itu.dk/research/cla/externals/clib.} The tests have been executed on a standard desktop computer (\emph{Intel\textregistered Core\texttrademark 2 Quad CPU Q9400} CPU with \emph{2.66GHz} and \emph{2GB} RAM) using the CLib library. We compared the performance of \textsc{Sequential} and \textsc{CoreDiag} for the different configuration knowledge bases. In order to show the advantages  of \emph{\textsc{CoreDiag}} that come along with an increasing number of redundant constraints, we generated three additional versions from the benchmark knowledge bases (see Table 1) that differ in their \emph{redundancy rate} $R$ (see Formula 1). The number of iterations per setting was set to 10; for each iteration we applied a randomized constraint ordering. Note that an evaluation of the individual properties of the used knowledge bases is within the scope of future work.

\begin{equation}
R(C_{KB})= \frac{|redundant ~constraints~ in ~C_{KB}|}{|constraints~ in~ C_{KB}|}
\end{equation}

In addition to the original version (redundancy rate = \textasciitilde 0-10\%) we generated three  knowledge bases with the redundancy rates 50\%, 75\%, and 87.5\%. For example, a knowledge base with redundancy rate 50\% can be generated by simply duplicating each constraint of the original knowledge base. Starting with a redundancy rate  of 50\% we can observe a transition in the runtime performance (\textsc{CoreDiag} starts to perform better than \textsc{Sequential})  due to the increased number of redundant constraints (see the \emph{large2} configuration knowledge base in Figure 1). Another outcome of our analysis is that nearly each of the investigated configuration knowledge bases contains redundant constraints (see Table 1). The average runtime for determining configurations without the redundant constraints is lower compared to the runtime with the redundant constraints included (see Table 2) -- for this evaluation as well the number of iterations per setting was set to 10; for each iteration we applied a randomized constraint ordering.

\section{Conclusions}
The detection of redundant constraints plays a major role in the context of (configuration) knowledge base development and maintenance. In this paper we have proposed two algorithms which can be applied for the identification of minimal cores, i.e., minimal sets of constraints that preserve the semantics of the original knowledge base. The \emph{\textsc{Sequential}} algorithm can be applied in settings where the number of redundant constraints in the knowledge base is  low. The second algorithm (\emph{\textsc{CoreDiag}}) is more efficient but restricted in its application to knowledge bases that contain a large number of redundant constraints.

\section{Acknowledgements}
The work presented in this paper has been conducted within the scope of the research project ICONE (Intelligent Assistance for Configuration Knowledge Base Development and Maintenance) funded by the Austrian Research Promotion Agency (827587).

\bibliographystyle{named}
\bibliography{references}

\end{document}